\documentclass{article}

\usepackage{arxiv}

\usepackage[utf8]{inputenc} 
\usepackage[T1]{fontenc}    
\usepackage{hyperref}       
\usepackage{url}            
\usepackage{booktabs}       
\usepackage{amsfonts}       
\usepackage{nicefrac}       
\usepackage{microtype}      
\usepackage{lipsum}
\usepackage{graphicx}
\graphicspath{ {./images/} }

\usepackage{cite}
\usepackage{amsmath,amssymb,amsfonts}
\usepackage{algorithmic}
\usepackage{graphicx}
\usepackage{textcomp}
\usepackage{float}
\usepackage{subcaption}

\title{On-site estimation of battery electrochemical parameters via transfer learning based physics-informed neural network approach}

\author{
 Josu Yeregui* \\
  Electronic and Computer Science Department\\
  Mondragon Unibertsitatea\\
  Mondragon Spain 20500.\\
  \texttt{jyeregui@mondragon.edu} \\
   \And
   Iker Lopetegi \\
  Electronic and Computer Science Department\\
  Mondragon Unibertsitatea\\
  Mondragon Spain 20500.\\
  \texttt{ilopetegui@mondragon.edu} \\
   \And
   Sergio Fernandez \\
  Electronic and Computer Science Department\\
  Mondragon Unibertsitatea\\
  Mondragon Spain 20500.\\
  \texttt{sfernandezg@mondragon.edu} \\
   \And
   Erik Garayalde \\
  Electronic and Computer Science Department\\
  Mondragon Unibertsitatea\\
  Mondragon Spain 20500.\\
  \texttt{egarayalde@mondragon.edu} \\
   \And
   Unai Iraola \\
  Electronic and Computer Science Department\\
  Mondragon Unibertsitatea\\
  Mondragon Spain 20500.\\
  \texttt{uiraola@mondragon.edu} \\
}

\begin{document}
\maketitle
\begin{abstract}
This paper presents a novel physical parameter estimation framework for on-site model characterization, using a two-phase modelling strategy with Physics-Informed Neural Networks (PINNs) and transfer learning (TL). In the first phase, a PINN is trained using only the physical principles of the single particle model (SPM) equations. In the second phase, the majority of the PINN parameters are frozen, while critical electrochemical parameters are set as trainable and adjusted using real-world voltage profile data. The proposed approach significantly reduces computational costs, making it suitable for real-time implementation on Battery Management Systems (BMS). Additionally, as the initial phase does not require field data, the model is easy to deploy with minimal setup requirements. With the proposed methodology, we have been able to effectively estimate relevant electrochemical parameters with operating data. This has been proved estimating diffusivities and active material volume fractions with charge data in different degradation conditions. The methodology is experimentally validated in a Raspberry Pi device using data from a standard charge profile with a 3.89\% relative accuracy estimating the active material volume fractions of a NMC cell with 82.09\% of its nominal capacity.
\end{abstract}


\section{Introduction}
\label{Sec:introduction}

Optimal battery control requires the estimation of many non-measurable variables and parameters, such as state of charge (SOC), state of health (SOH) or remaining useful life (RUL) among others. Empirical models that leverage on experimentally derived relationships \cite{Yuanyuan} or data-driven techniques to extract patterns from historical data \cite{Che} have shown good performance for these tasks, provided good quality data is available. Nevertheless, these approaches often lack interpretability, as they do not capture the underlying internal processes driving battery behaviour. As a result, they fail to provide insight about the cell state that could be used for optimal control or to prevent degradation under varying operating conditions.

In view of the above, there is a push to adopt methodologies considering physics-based models (PBM), as they help to better understand how cell performance is affected by different degradation mechanisms and modes \cite{Edge}. State estimators coupled with PBMs provide momentary electrode-specific state of health indicators alongside the internal physical state of the battery \cite{Lopetegi_2024}. Even if the information gained by the physics-based methodologies can enable advanced control and prognostics, they rely on the fundamental model to be adequately parameterized. Since the initial set of parameters may vary between cells of different batches, and will evolve over the battery's lifecycle, dynamically adjusting and correcting the electrochemical parameters during operation remains a key challenge. In this regard, it is feasible to group key parameters and update them using measured voltage under controlled conditions \cite{FAN2023232555}. Yet, such strategies require performing specific tests, leaving the battery unusable during the systematic update, and suffer from potentially time-intensive simulations. Consequently, the parameterization issue hinders the implementability of PBMs, and are often deployed and validated offline.

Recently, methods to unify the adaptability of data-driven models with the interpretability of PBMs have been proposed \cite{Perspective}. Among these, PINNs stand out due to their ability to incorporate physical laws directly into the learning process \cite{raissiI, raissiII}. By embedding differential equations into the loss function, PINNs can efficiently solve both forward and inverse problems, making them especially effective for parameter identification and scenarios with limited data availability. PINNs are particularly useful, as they are suitable for parameter identification by solving the inverse problem, as the interest on other research areas indicate. Interestingly, Xu et.al. \cite{Xu} propose a TL-based PINN framework for inverse problems in engineering. It integrates physics constraints and multi-task learning to estimate unknown loads from limited displacement data. A two-stage process, offline pre-training and online fine-tuning, enables them adapting to new structures and loading conditions while reducing data and computational costs. While some works have attempted the use of PINNs for battery SOH with great accuracy and robustness \cite{Wang2024}, they mostly rely on empirical principles, capturing observed behavioural capacity fading patterns identified on battery systems. Despite this, models that incorporate the internal dynamics of the cell like the ones PBM represent are being developed. Singh et.al. \cite{Singh} present a PINN model where the forward problem is solved to estimate the momentary battery state and infer the SOC and SOH. Focusing on the parameter identification,  Zheng et.al. \cite{Zheng} build two physics-informed DeepONets that predict Li-ion concentration in the anode and cathode based on the current function and solid diffusion coefficients, followed by a third DeepONet that predicts terminal voltage from the surface concentration values and diffusion coefficients. The model updates the anode diffusion coefficient and a parameter ($\beta$) governing the current function by gradient descent, this last one validating that the model converges to an optimal value. However, the inferred parameters do not address ageing conditions. Also, the computational cost can grow up significantly for each parameter estimation that has to be done, adding trainable parameters in the branch net and requiring the whole model's training, limiting the implementation capabilities.

This paper presents a two-phase physical parameter estimation framework using TL. In the first phase, a PINN is trained only using physical knowledge of a lithium-ion battery PBM, so that the dynamics under different input current scenarios are learnt. In the second phase, the majority of the PINN parameters are frozen while critical physical parameters are set as trainable, adapting them using voltage profile data obtained from the field application. Due to the TL strategy, the deployed model has very low computational cost, opening the possibility to implement the strategy on a BMS. Additionally, because the first phase requires no field data, the model requires little input to set up and work as intended. Summing up, the main contributions of this paper are: 

\begin{itemize}
    \item Proposed a method to adapt PBM parameters using operation data.
    \item Proposed an online physical parameter estimation approach that only requires operating data.
    \item The parameter estimation strategy can be implemented in an edge device.
    \item Approach validated in simulated and experimental setups under different ageing conditions.
\end{itemize}

\section{Methodology}\label{Sec:Methodology}

\subsection{Single Particle Model}\label{Sec2:SPM}


We will gather the underlying physical fundamentals from well validated PBM methodologies for our battery application oriented PINN. Among the PBMs that can be found, the Single Particle Model (SPM) simplifies the extensively used P2D model making it easier to implement and reducing its computational cost significantly, despite their known issues working with high discharge/charge C-rates \cite{SPMfirst}. The SPM considers the solid-phase diffusion to be the most dominant dynamic inside the cell. This way, electrolyte dynamics are neglected and the reaction rate is assumed uniform for each electrode. Therefore, the electrodes can be described solving the dynamics of an average solid phase spherical particle for each electrode. This simplification makes the system dependant just on the temporal domain and one spatial coordinate system, and contains an output function just dependant on the solid phase dynamics. Following the formulation of the SPM in \cite{SPMBizeray}, the lithium concentration in the active material is described by the Fickian diffusion equation in spherical coordinates:
\begin{equation}\label{eq.fick}
  \frac{{\partial {\bar{c}_i}}}{{\partial t}} = \frac{{{1}}}{{{\bar{r}_i^2}}}\frac{\partial }{{\partial r_i}}\left( {{D_i \bar{r}_i^2}\frac{{\partial {\bar{c}_i}}}{{\partial \bar{r}_i}}} \right) \quad \text{for } i \in \{p, n\},  
\end{equation}

\noindent where $p$ and $n$ are the subscripts representing the cathode and anode respectively, $\bar{r}_i$ is the radial coordinate, $\bar{c}_i$ the active material concentration, and $D_i$ is the diffusion coefficient in each electrode $i$. We assume the $D_i$ to be constant and uniform along the particles and w.r.t. concentration. 

As mentioned, the model primarily depends on the solid phase concentration, which will take values in different ranges depending on the cell characteristics. However, it is possible to reformulate the model using the maximum concentration of the electrodes, $c^{max}_i$, so that we can work with the stoichiometry in the governing equations. This stoichiometry $c_i = \bar{c}_i/c^{max}_i$ ranges between 0 and 1, and can be adopted in all model equations. Similarly, the radial coordinates can be redefined as a dimensionless parameter by $r = \bar{r}_i/R_i$, a parameter in the same range. This way, we define the dimensionless SPM model:

\begin{align}\label{eq.fickDim}
\frac{{\partial {c_i}}}{{\partial t}} = \frac{{D_i}}{{{R_i^2\bar{r}^2}}}\frac{\partial }{{\partial \bar{r}}}\left( {{r^2}\frac{{\partial {c_i}}}{{\partial r}}} \right) \quad \text{for } i \in \{p, n\}.  
\end{align}

The initial condition of concentration is first contemplated:
\begin{equation}\label{eq.ivcond}
  c_i(0, r_i) = c_{i,0} \quad \text{for } i \in \{p, n\},  
\end{equation}

\noindent where $c_{i,0}$ is the initial concentration profile in each particle. The diffusion \eqref{eq.fickDim} is also subject to Neumann boundary conditions at the centres $r_i = 0$ and surfaces $r_i = R_i$ of both particles:
\begin{align}
    \frac{{\partial {c_i}}}{{\partial r_i}} \bigg\rvert_{r_i = 0} &= 0 , \label{eq.bcccond}\\ 
    D_i \frac{{\partial {c_i}}}{{\partial r_i}} \bigg\rvert_{r_i = 1}&= - \frac{R_i}{c^{max}_i} j_i \quad \text{for } i \in \{p, n\} \label{eq.bcscond}.
\end{align}

Here, $j_i$ represents the reaction rate or flux of lithium out of the particle. In the SPM the reaction rate can be directly related to the current $I$ applied to the cell in the following way:

\begin{align}\label{eq.flux}
    j_p = \frac{-I}{a_p \delta_p F A} \qquad\text{and}\qquad j_n = \frac{I}{a_n \delta_n F A},
\end{align}
\noindent where $a_i$ is the surface area of the active material given by $a_i = 3\epsilon/R_i$, $\epsilon$ is the volume fraction of the electrode's active material, $\delta_i$ is the thickness of the electrode, $F$ is Faraday’s constant, and $A$ is the total surface area of the electrode that we assume to be equal for the cathode and anode.


Solving the concentration, we can define the battery terminal voltage, $V$, as the sum of the open circuit potentials (OCPs), $U^{ocp}_i$, and the overpotentials , $\eta_i$, from both electrodes:
\begin{align}\label{eq.VSPM}
    V = U^{ocp}_p(c^s_p) - U^{ocp}_n(c^s_n) + \eta_p - \eta_n.
\end{align}

This lets us obtain the voltage variable that we can measure from the real cell easily. The OCP is defined with a function of the surface concentration, $c^s_i$, which we obtain solving the PDE in \eqref{eq.fickDim}. The overpotential relates to the flux $j_i$ by the Butler-Volmer kinetics equation \cite{Doyle}. For this model, we assume the electrolyte concentration $c_e$ to be constant, and the  anodic and cathodic charge transfer coefficients equal. Hence, the overpotentials can be computed by:
\begin{align}
    \eta_i &= \frac{2RT}{F} \sinh^{-1}{\left(\frac{j_i F}{2 i_{0,i}}\right)} , \label{eq.overpot}\\
    i_{0,i} &= k_i F \sqrt{c_e} \sqrt{c^s_i} \sqrt{c^{max}_i - c^s_i} \quad \text{for } i \in \{p, n\}, \label{eq.transfer}
\end{align}

\noindent where $R$ is the universal gas constant and $T$ the average cell temperature. The exchange current density $i_{0,i}$ depends on the reaction rate constant $k_i$, the surface and maximum concentrations ($c^s_i$ and $c^{max}_i$, respectively) of the electrode, and the aforementioned constant electrolyte concentration, $c_e$.

As we can see in this model, the simulated behaviour of the battery relies on specific physical parameters. On the one hand, we contemplate geometrical factors such as particle size $R_i$, electrode area $A$, and thickness $\delta_i$, to remain fixed since the battery's assembly. On the other hand, we can expect other parameters, such as the diffusion coefficients $D_i$, and active material volume fractions $\epsilon_i$, to evolve over time due to factors like ageing and chemical side reactions.

\subsection{Proposed transfer learning based PINN approach}\label{Sec2:Approach}

\begin{figure}
    \centering
    \includegraphics[width=0.45\textwidth]{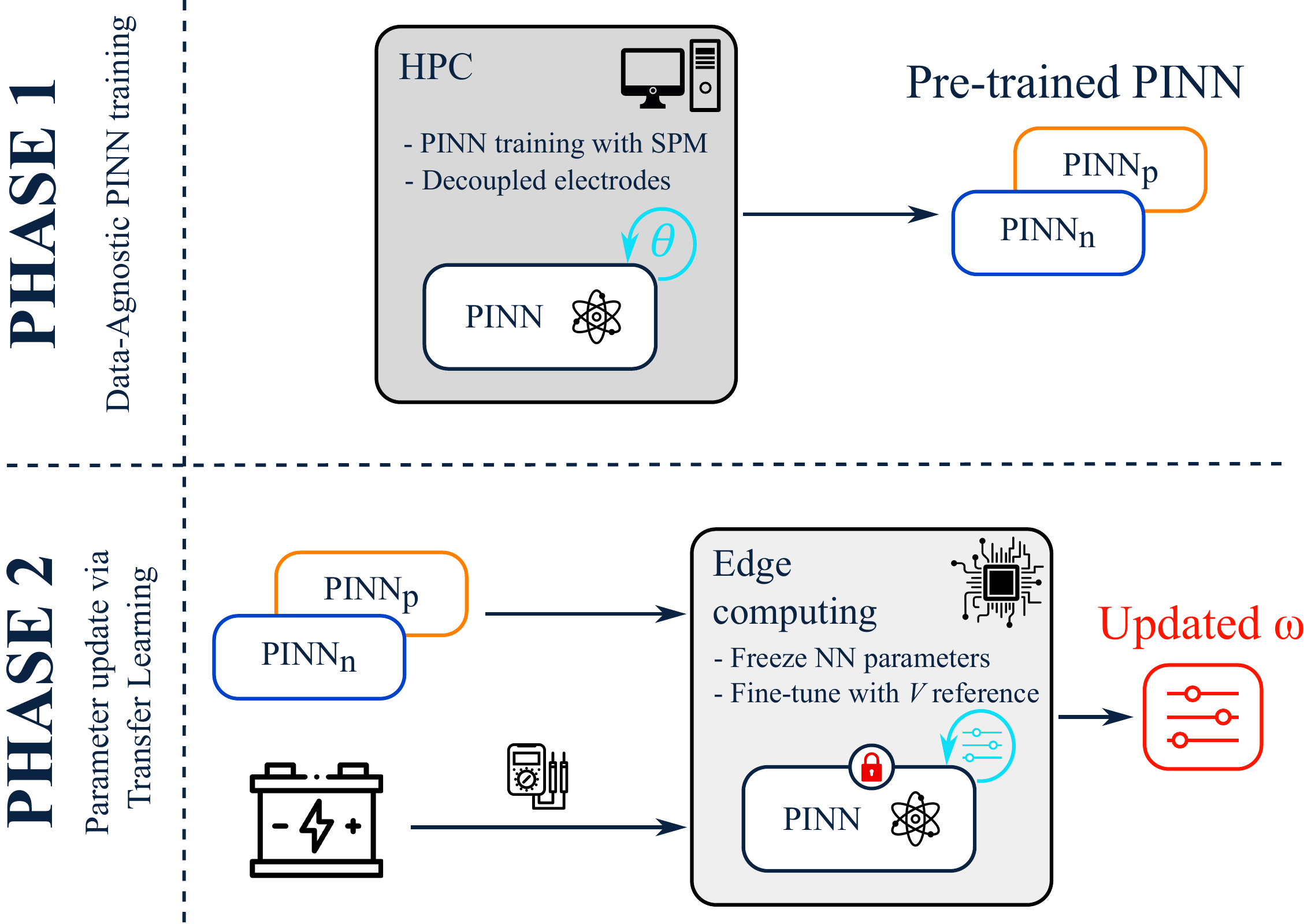}
    \caption{Two-phase methodology solution proposed in this work. During Phase 1 the PINNs for both electrodes are trained without external measurements. In Phase 2 selected electrochemical parameters $\omega$ from the obtained voltage profile.}
    \label{fig:Methodology}
\end{figure}

For our proposed solution, we adopt the PINN modelling strategy with the end goal of estimating the battery's representative physical parameters during its operating cycle. We split the methodology so that the PINN is pre-trained in a laboratory environment in a first phase, and a transfer learning strategy is applied in the second phase to adjust the physical parameters to the cell in the application in a computationally efficient way (Figure \ref{fig:Methodology}).

\subsubsection{Phase 1: Data-agnostic PINN training}

\begin{figure*}
    \centering
    \includegraphics[width=\textwidth]{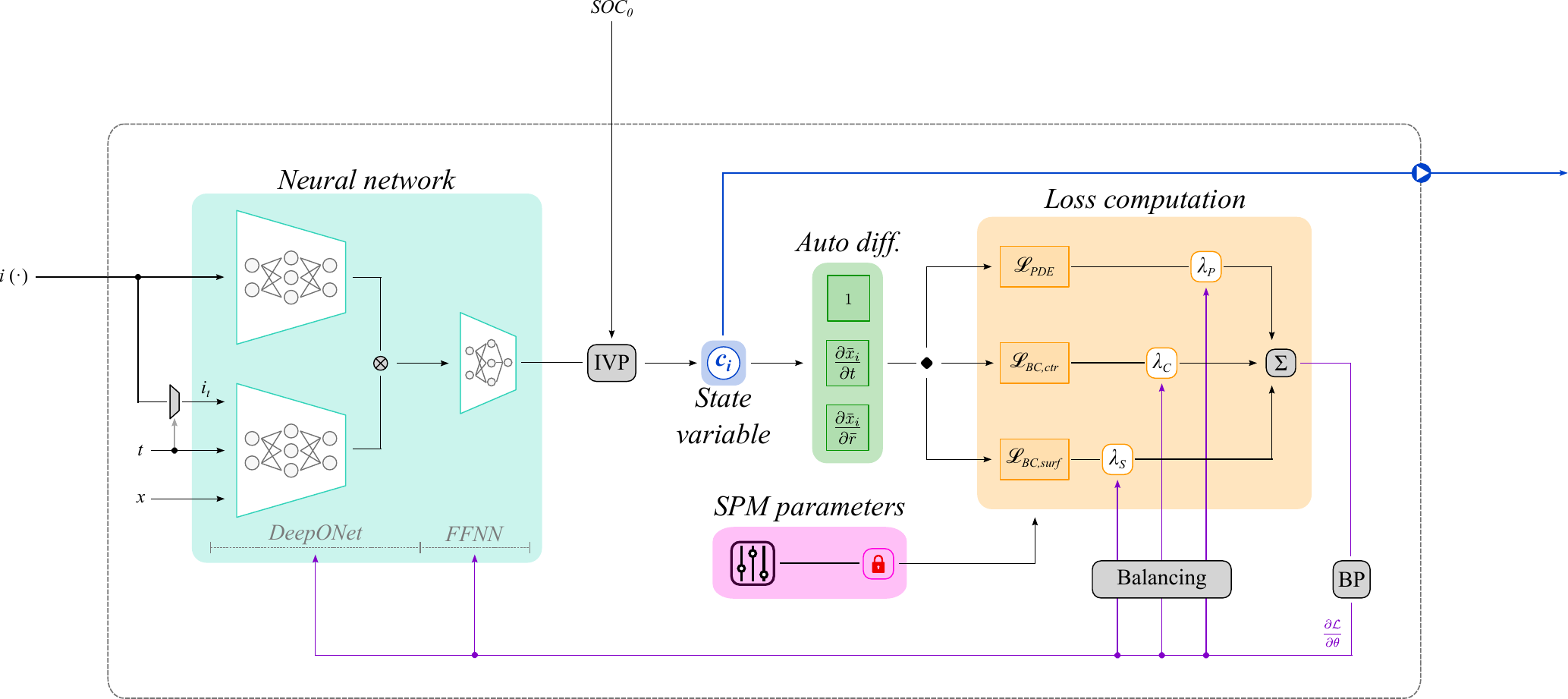}
    \caption{Phase 1 PINN schematic for the negative electrode. Input current profiles and initial $SOC_0$ values can be synthetically produced for the training/validation. The trained model outputs the concentration values for a given test condition.}
    \label{fig:Phase1}
\end{figure*}

At a first part of the methodology, the PINN model is defined and trained through simulation-driven methodologies. Being based on a PINN, the model consists mainly on a NN and the physics-assisted loss function terms. This architecture is summarized in Figure \ref{fig:Phase1}. The NN outputs the dependant variable, lithium concentration $c_{\theta, i}$ at a given time $t$, particle collocation point $r$ and operating current $i_t$, and compares its predictive accuracy through the residuals from the physical equation. In our case study, the PINN borrows the underlaying physical equations from the SPM, so it is trained to replicate its behaviour. With a given set of physical parameters, which can be extracted from beginning of life experimentation or literature, the model should be able to be trained to describe the SPM behaviour.

For the NN part of the system, we have chosen to combine a DeepONet architecture with dense layers before the output. The DeepONet splits the network in a branch network that processes input functions (current function $i(\cdot)$) and a trunk network that processes collocation points (time $t$ and radius point $\overline{r}$). Additionally, we feed the momentary current $i_t$ to the trunk network as extra information, like shown in Figure \ref{fig:Phase1}. This architecture is suited for capturing complex mappings between function spaces, making it highly effective for solving differential equations \cite{lu2021learning}. 

Our PINN approach is built with decoupled positive and negative electrodes to simplify the learning problem. The NN of each electrode is treated with the losses corresponding to their domain SPM equations. By treating each electrode separately, the PINN can independently learn the dynamics of each domain. This way, it is possible to study the mechanics such as diffusion characteristics, and ageing mechanisms in a separate way. Another benefit is the flexibility in refining the model for each electrode, avoiding the propagation of high error cases among both domains.

Next, we need to define the loss function to train the PINN. In this first phase, we want the model to replicate the behaviour of the underlying SPM model. Based on the SPM principles described in \eqref{eq.fickDim}-\eqref{eq.bcscond}, the total loss will be the sum of the terms that describe the battery model. Thus, the terms corresponding to the PDE $\mathcal{L}_{PDE}$, particle centre boundary condition $\mathcal{L}_{BCc}$, particle surface boundary condition $\mathcal{L}_{BCs}$, and initial value $\mathcal{L}_{IV}$ are added:
\begin{equation}\label{eq.PINN_loss}
\mathcal{L} = \mathcal{L}_{PDE} + \mathcal{L}_{BCc} + \mathcal{L}_{BCs} + \mathcal{L}_{IV}\; ,
\end{equation}

\noindent where
\begin{align}
\mathcal{L}_{\text{PDE}} &= \frac{1}{N_{\text{PDE}}} \sum_{i=1}^{N_{\text{PDE}}} \left\| \frac{{\partial {c_{\theta, i}}}}{{\partial t}} - \frac{{D_i}}{{{R_i^2\bar{r}^2}}}\frac{\partial }{{\partial \bar{r}}}\left( {{\bar{r}^2}\frac{{\partial {c_{\theta, i}}}}{{\partial \bar{r}}}} \right) \right\| , \label{eq.PDE} \\ 
\mathcal{L}_{\text{BCc}} &= \frac{1}{N_{\text{BCc}}} \sum_{i=1}^{N_{\text{BCc}}} \left\| \frac{{\partial {c_{\theta, i}}}}{{\partial \bar{r}_i}} \right\| , \label{eq.BCc} \\
\mathcal{L}_{\text{BCs}} &= \frac{1}{N_{\text{BCs}}} \sum_{i=1}^{N_{\text{BCs}}} \left\| D_i \frac{{\partial {c_{\theta, i}}}}{{\partial \bar{r}_i}} + \frac{R_i}{c^{\max}_i} j_i \right\| , \label{eq.BCs} \\
\mathcal{L}_{\text{IV}} &= \frac{1}{N_{\text{IV}}} \sum_{i=1}^{N_{\text{IV}}} \left\| c_{\theta, i}(0, r_i) - \bar{c}_{i,0} \right\| , \quad \text{for } i \in \{p, n\}. \label{eq.IV}
\end{align}

\noindent Note that all these terms can be computed using just the NN output and its gradients w.r.t. the independent variables. Therefore, no measurement data from the application is needed to update the NN parameters, so we call the learning of this phase ``data-agnostic".

The PDE points are chosen with a random or grid-uniform distribution along the time-space domain for each electrode. In this work, a Hammersley quasi-random method has been employed in that 2D space to reduce the discrepancy. Here, the first domain dimension uses a uniform sequence, while the rest of dimensions are filled reversing the base n (base 2 for this work) representation of the former sequence. Despite this being a deterministic sampling method, the method ensures an evenly spaced distribution without growing the number of points exponentially like in a grid sampling, which affects computational cost; and avoiding clusters of points that may happen in random methods, reducing the risk of overfitting.

Having four different terms to be considered by the loss function, we need to ensure that all terms are properly minimised during training. Since the prior solution to the problem is unknown, we can expect the terms to be different both in scale and response to the gradient descent optimisation. Unfortunately, it is not possible to normalise losses before training. Hence, the most common way to deal with this balancing problem is to weight each term by a manually defined constant value before training. This is problematic because it adds additional hyperparameters to tune, and we will not be able to ensure a proper generalisation for different cases of input functions.

To solve this issue, we have implemented a self-adaptive weighting strategy, so that all losses are automatically balanced during the training process. We follow the balancing method in \cite{Wang2023AnEG}, which consists on equalising the gradients of each weighted loss term to each other. In this framework, each term's weights $\lambda$ relate to the gradients in the following way:
\begin{equation}\label{eq.PINN_lossbalance}
\begin{split}
\|\lambda_{i}\nabla_\theta \mathcal{L}_{i} \| 
&= \sum_i \|\nabla_\theta \mathcal{L}_{i} \| \\
&\quad \text{for } i \in \{\text{PDE, BCc, BCs, IV}\},
\end{split}
\end{equation}

\noindent so new $\hat{\lambda}$ values can be obtained after each training step. However, performing an update after each iteration can be unstable, so $\lambda_{k+1}$ is computed with the running average of the previous weights every 10 iterations, modulated by a user defined $\alpha$ parameter.
\begin{align}
\hat{\lambda}_{i} &= \frac{\sum_{i} \|\nabla_\theta \mathcal{L}_{i} \|}{\|\nabla_\theta \mathcal{L}_{i} \|}, \label{eq.PINN_lambdacomp} \\
\begin{split}
\lambda_{k+1, i} &= \alpha \cdot \lambda_{k, i} + (1 - \alpha) \cdot \hat{\lambda}_{i} \\
&\quad \text{for } i \in \{\text{PDE, BCc, BCs, IV}\} . \label{eq.PINN_lambdaMA} 
\end{split}
\end{align}

\noindent This way, the total loss function described in \eqref{eq.PINN_loss} is reformulated to:
\begin{equation}\label{eq.PINN_loss_weight}
\mathcal{L} = \sum_{i} \lambda_{k, i} \mathcal{L}_{i}; \quad \text{for } i \in \{\text{PDE, BCc, BCs, IV}\},
\end{equation}

A final modification we have contemplated to favour the training process has been to attempt hard constraining the initial values and boundary conditions. This serves the purpose of reducing the number of terms in the added loss function and improve the balancing among the remaining terms. To achieve this, we transform the output of the NN so the initial value (IV) or boundary conditions (BC) can exactly match the values given by their equations without having to learn that behaviour. Unfortunately, both boundary conditions are Neumann BCs. Despite being feasible to apply similar transformation strategies for periodic domains, the battery model does not present such behaviour, and BCs have to be included on the loss function. Consequently, we modify \ref{eq.PINN_loss_weight} to just include the PDE, BCc and BCs terms.

Conversely, the initial value can be imposed by the transformation given in \cite{chen2020neurodiffeq}, as we expect it to be a scalar function. Given the initial condition of the concentration, the NN architecture can be adapted, so that:
\begin{equation}\label{eq.PINN_IVhard}
c_{\theta, i}(t, r) = (1-e^{\beta t}) \cdot c_{NN}(t, r) + c_{0,k} \quad \text{for } i \in \{p, n\},
\end{equation}

\noindent with $c_{NN}$ being the output of the NN and $\beta$ a constant decay factor. $c_{0,k}$ is computed taking into account the maximum, $c_{max}$, and minimum, $c_{min}$, concentrations of each electrode, and the input initial state of charge $SOC_0$:
\begin{equation}\label{eq.C_ini}
c_{0,k} = \frac{SOC_0 - \frac{c_{min, k}}{c_{max, k}}}{1 - \frac{c_{min, k}}{c_{max, k}}} \quad \text{for } i \in \{p, n\}.
\end{equation}

Because of this transformation, we can ensure the output at $t = 0$ to be exactly $c_{i,0}$ in both electrodes, so the loss term in \eqref{eq.IV} becomes redundant. Thus, we remove it from the loss computation on \eqref{eq.PINN_loss_weight}, reducing the number of terms and weight factors to three, like shown in Figure \ref{fig:Phase1}.

\subsubsection{Phase 2: Fine-tuning for parameter exploration}

The PINNs for each electrode trained in the first phase provide us a model that behaves according to the introduced set of physical parameters. This results in a model that accurately reflects the behaviour of the electrodes under the specified conditions, effectively embedding the physics of the problem into the neural network. 

\begin{figure*}
    \centering
    \includegraphics[width=\textwidth]{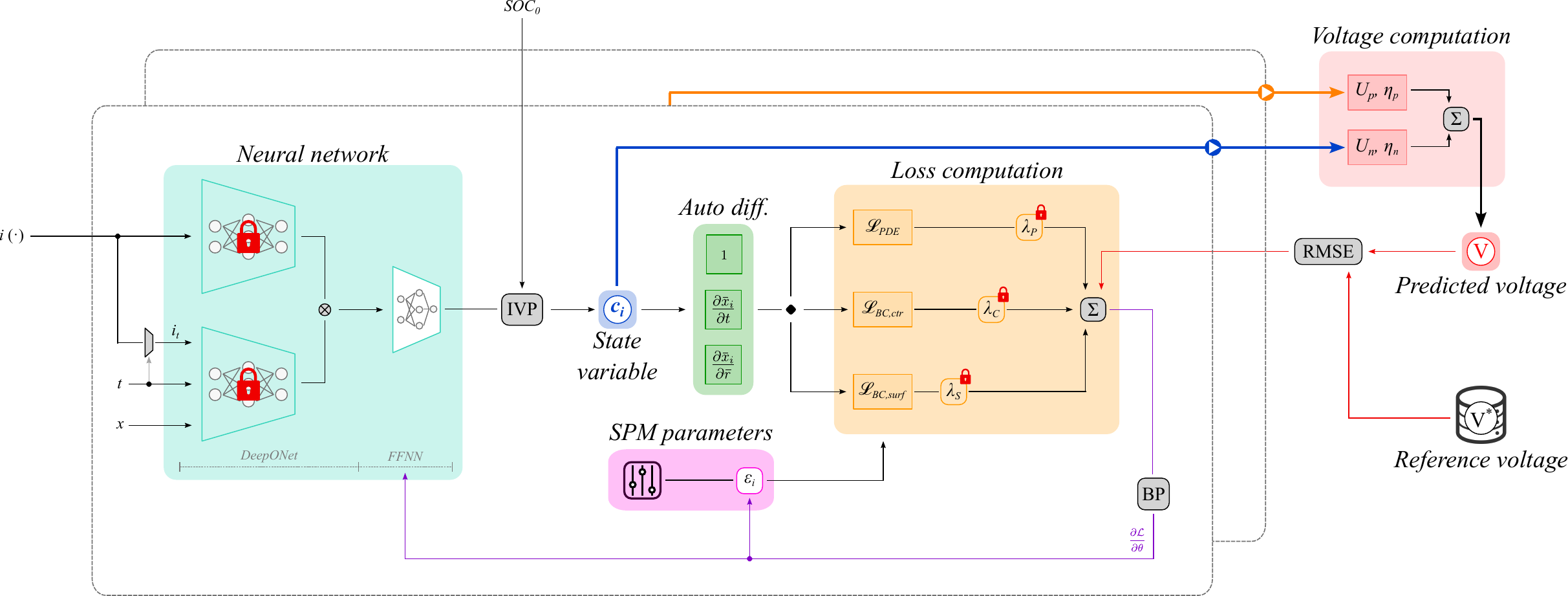}
    \caption{Phase 2 PINN model schematic. Concentrations from positive and negative electrodes are used to compute the terminal voltage estimate. DeepONet parameters and adaptable loss weights are frozen while enabling the learning of some physical parameters by gradient descent.}
    \label{fig:Phase2}
\end{figure*}

In the second phase, we apply a transfer learning strategy to adjust the response of the PINN to the empirical behaviour based on the measurements we can take from a battery in an application. In this regard, the model's output concentrations will not be available, so we compute the measurable terminal voltage $V$ from both electrodes using SPM's equations \eqref{eq.VSPM}-\eqref{eq.transfer}. This phase's methodology with the added voltage computation is shown in Figure \ref{fig:Phase2}.

While we keep the loss terms corresponding the SPM retraining both electrode PINNs, a new shared term is added to their corresponding loss functions defined in \eqref{eq.PINN_loss_weight}. This term is the error between the predicted terminal voltage by the PINN's $V$ and the measured voltage $V^*$:
\begin{align}
\mathcal{L}_{\text{V}} &= \frac{1}{N_{\text{V}}} \sum_{i=1}^{N_{\text{V}}} \left\| V_i - V^*_i \right\| . \label{eq.Vloss} 
\end{align}


This way, the model will respect the physical fundamentals while fine-tuning for the measured behaviour. At this stage, the previously trained PINN's DeepONet part is frozen, whose layers will account for the learned electrochemical principles. The FFNN is left as trainable to consider the new dynamics when added the voltage loss term. Setting a great part of the learnable parameters in the DeepONet, we reduce the computational cost of the fine-tuning process in this phase significantly, potentially enabling it for edge implementation.


Under this arrangement, we can expect the SPM related loss functions to worsen if the measured voltage differs significantly from the PINN's calculated one. The reason for this is that the physical parameter set does not correspond to the application cell, because of volatile factors in the manufacturing process or ageing reasons. Accordingly, we set some selected physical parameters as learnable parameters in this phase, potentially uncovering system-specific behaviours that were not explicitly defined in the initial set of parameters. These are initialized with the preset values (defined for Phase 1) or a previous checkpoint from executing this phase, and evolve alongside the FFNN parameters of the PINN.

Like we mentioned when defining the SPM, parameters such as solid-phase diffusions $D_i$ and active material volume fractions $\epsilon$ are more susceptible to change from a definition on a laboratory environment or ageing conditions. Stoichiometric limits, described with the maximum and minimum concentrations $c^{max/min}_i$, can also be affected by these reasons. Active material volume fractions $\epsilon$ for both electrodes are particularly interesting, as they are related to the Loss of Active Material (LAM) values of both electrodes. These can be experimentally estimated at certain points during the lifecycle of the cell with some optimizations after performing specific check-up cycles \cite{Sergi}.

With the proposed methodology, we can obtain a theoretical Beginning of Life (BOL) cell PINN model in Phase 1 and fine-tune it at low computational cost with experimental voltage data in Phase 2, outputting a selected number of evolved physical parameters during its lifecycle. 

\section{Simulation results}\label{Sec:Results}

\subsection{Setup}\label{Sec3:Setup}


In this section, we apply the proposed two-phase methodology, training the model in a local machine and subject it to an operating condition. To facilitate the deployment, the reference battery is fully simulated environment using the PyBaMM (Python Battery Mathematical Modeling) framework \cite{Sulzer2021}. This way, Phase 1 PINN model's validity will be tested against the PyBaMM SPM battery, and the field data needed to fine-tune in Phase 2 will be acquired from this environment. By simulating the internal states of a battery with this tool we allow a more in-depth comparison and validation of the model in both phases. The electrochemical parameters used in the simulations follow the LG M50 parameter set provided by \cite{Chen_2020}, loaded as a default PyBaMM parameter set. The needed parameters for the SPM computation are shown in the supplementary material S1.

The PINN implementation has been done under the PyTorch framework, using its \textit{autograd} functionality for automatic differentiation. The needed computations, including the PBM model executions have been carried out in a local machine with an Intel\textsuperscript{\tiny\textregistered} Core\texttrademark{} i5-12400 CPU, 32GB of RAM and a NVIDIA\textsuperscript{\tiny\textregistered} GeForce\textsuperscript{\tiny\textregistered} RTX 3060 GPU.

\subsection{Phase 1: Pre-training of the model}\label{Sec3:Pretraining}

For the first phase, we set up a DeepONet architecture with the configuration described in the methodology. All hyperparameters and implementation details have been gathered in the supplementary material S2. The model is trained with different C-rates on full charge/discharge conditions, randomly chosen after each epoch. We reserve specific cycling conditions for validation, showing a 1C charge rate working condition as the final test set.

With this setup, the PINN model for each electrode has been trained over 50000 epoch. Trained in the local machine, it has taken an average of 95.23 ms for each training iteration, with a total computational cost of 1 hour 19 minutes to complete the training of both models. The procedure has resulted on a 2.110e-4 RMSE total loss for the positive electrode and 1.743e-4 RMSE total loss for the negative electrode in the 1C charge test condition.

\begin{figure*}
    \centering
    \subfloat[\centering\label{fig: eps_p09n1}]{{\includegraphics[width=0.45\textwidth]{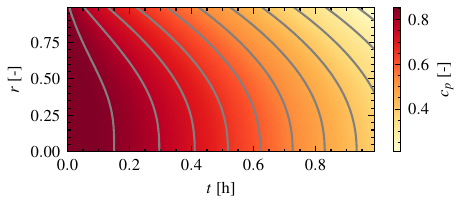}}}\qquad
    \subfloat[\centering\label{fig: V_p09n1}]{{\includegraphics[width=0.45\textwidth]{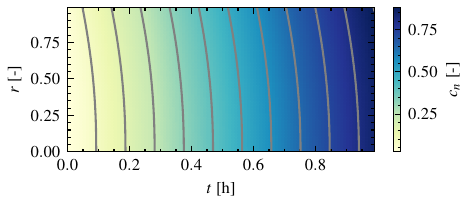}}}\\
    \subfloat[\centering\label{fig: eps_p09n1}]{{\includegraphics[width=0.45\textwidth]{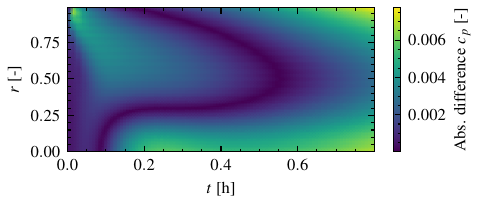}}}\qquad
    \subfloat[\centering\label{fig: V_p09n1}]{{\includegraphics[width=0.45\textwidth]{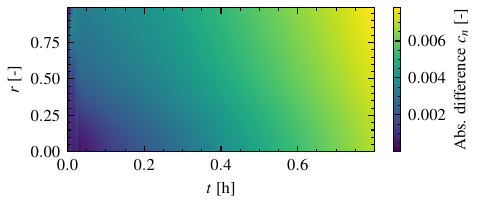}}}\\
    \caption{Predicted concentration profiles compared to the PyBaMM reference model for the 1C charge condition. Up: Concentration profile in the (a) positive electrode and (b) negative electrode. Down: absolute difference between the predicted and reference concentration in the (c) positive electrode and (d) negative electrode.}
    \label{fig: concentrations}
\end{figure*}

We can evaluate the learned performances by studying the predicted concentrations against the reference model. Figure \ref{fig: concentrations} shows the concentration profiles for the 1C charge condition in both electrodes after the training. We compare these concentrations to the PyBaMM reference model, as well as the residual in the full test dimension obtained from \eqref{eq.fickDim}. In general, the model demonstrates a strong ability to predict the concentration accurately across the spatial and temporal domains, even in the absence of field data. This results in a model behaving similar to the PyBaMM reference model, with a total of 2.835e-3 RMSE in the positive electrode and 4.883e-3 RMSE in the negative. The major discrepancies, both in error and residual difference, are found close to the surface boundary at the beginning of the test. We can expect this behaviour, as the most significant gradients will be found in that region with CC tests.

\begin{figure}
    \centering
    \includegraphics{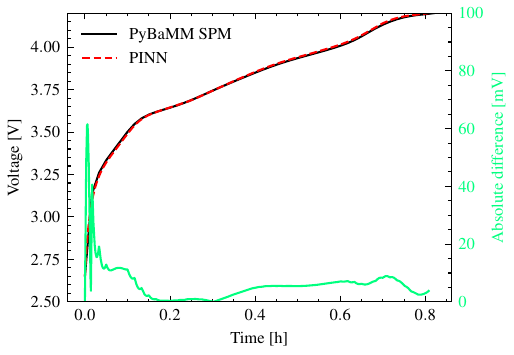}
    \caption{Reconstructed voltage response using the trained electrode models compared to the PyBaMM reference model for the 1C charge condition, with their difference in mV.}
    \label{fig: V_pretrain}
\end{figure}

Despite this, the concentration prediction is accurate enough to replicate the simulated model behaviour. We can check this fact by obtaining the output concentrations of both trained electrodes and reconstructing the voltage using \eqref{eq.VSPM}-\eqref{eq.transfer}, like we will do in Phase 2. This provides a RMSE of 8.94 mV against the PyBaMM model response at the 1C test, shown in Figure \ref{fig: V_pretrain}.

\subsection{Phase 2: Parameter estimation}\label{Sec3:Finetune}


Both positive and negative electrode PINNs have been trained independently, each capturing the dynamics of their battery domain. In this phase, we  freeze their DeepONet parameters, alongside the loss weights before fine-tuning to the new behaviour. As previously mentioned, we set the ageing related $\epsilon$ parameters as trainable for both electrodes, so we can compute the LAM evolution on each electrode, and are initialized with their literature values. The output FFNN layers' parameters will also remain trainable, as the battery's behaviour is expected to change with the electrochemical parameters, adjusting the response concentrations for the same inputs in the new working conditions. We use Adam optimizer, fixing the learning parameters for both $\epsilon$ at 0.001 times their initial value ($\eta_{lr,i} = 0.001\cdot\epsilon_i$). For this process the same number of points as the Phase 1 training are evaluated on each iteration.


The voltage references are obtained from a simulated environment, changing the electrochemical parameters from literature and running the simulated behaviour in PyBaMM. This allows us to perform this phase's parameter updating in different battery conditions without the need to provoke each ageing scenario experimentally. As this method is aimed for deployment in application, we obtain a CC charging data from 0 $SOC$ to 4.2 V, which is common in many real scenarios. We perform the fine-tuning under the unseen 1C charge condition, used for the model performance verification. 

\begin{figure*}
    \centering
    \subfloat[\centering\label{fig: eps_p09n1}]{{\includegraphics[width=0.3\textwidth]{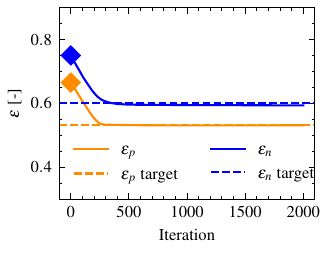}}}\qquad
    \subfloat[\centering\label{fig: eps_p09n1}]{{\includegraphics[width=0.3\textwidth]{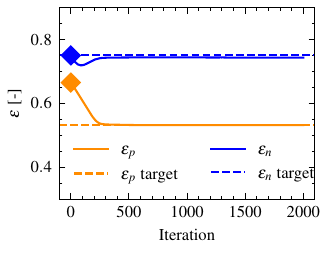}}}\qquad
    \subfloat[\centering\label{fig: eps_p09n1}]{{\includegraphics[width=0.3\textwidth]{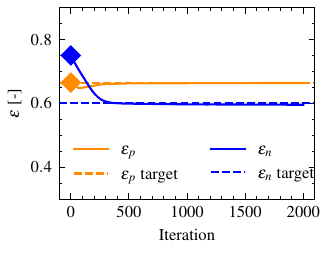}}}\\
    \subfloat[\centering\label{fig: eps_p09n1}]{{\includegraphics[width=0.3\textwidth]{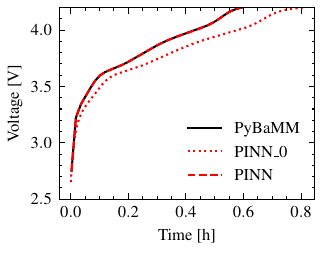}}}\qquad
    \subfloat[\centering\label{fig: eps_p09n1}]{{\includegraphics[width=0.3\textwidth]{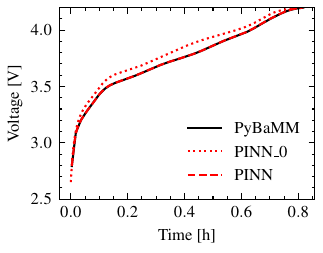}}}\qquad
    \subfloat[\centering\label{fig: eps_p09n1}]{{\includegraphics[width=0.3\textwidth]{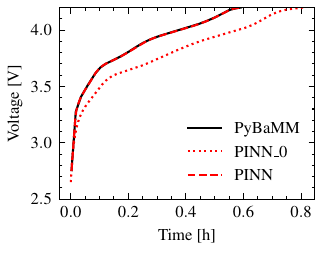}}}
    \caption{ Fine-tuning process for different active material volume fraction changes. (a) Evolution of $\epsilon_p$ and $\epsilon_n$ during fine-tuning when both parameters decrease ($\epsilon_p = 0.8$, $\epsilon_n = 0.8$). (b) Evolution when only $\epsilon_p$ decreases ($\epsilon_p = 0.8$, $\epsilon_n = 1.0$). (c) Evolution when only $\epsilon_n$ decreases ($\epsilon_p = 1.0$, $\epsilon_n = 0.8$). (d) Voltage prediction improvement during fine-tuning for case (a). (e) Voltage prediction improvement during fine-tuning for case (b). (f) Voltage prediction improvement during fine-tuning for case (c).}
    \label{fig: analytic}
\end{figure*}

Under this setup, we get voltage measurements changing $\epsilon$ in factors from 0.8 to 1. We limit the simulations to these conditions, as outside the bounds PyBaMM struggles to provide reliable responses. Based on observations, we have completed 2000 fine-tuning iterations for better result representation, but an early stopping strategy can be adopted looking at the parameters convergence when implementing the model. Figure \ref{fig: analytic} showcases the new learned behaviour by showing the initial PINN voltage prediction and it's approximation to the reference curve alongside $\epsilon_p$ and $\epsilon_n$ evolution during the fine-tuning; repeated for the case where both parameters change, or one of them changes.

Following this procedure, we have tried this methodology in the full space of the proposed use case, testing the estimation performance for different changes in the $\epsilon$ value for both electrodes. Here, we can relate the change in $\epsilon$ to the $LAM$ of each electrode by:

\begin{align}
LAM &= 1 - \frac{Q_{i}}{Q_{i, 0}},\\ 
Q_{i} &= F A L \epsilon_i c^{max}_{i}\left(\theta_{i, 100} - \theta_{i, 0} \right). \label{eq.LAM}
\end{align}

\begin{figure}
    \centering
    \includegraphics{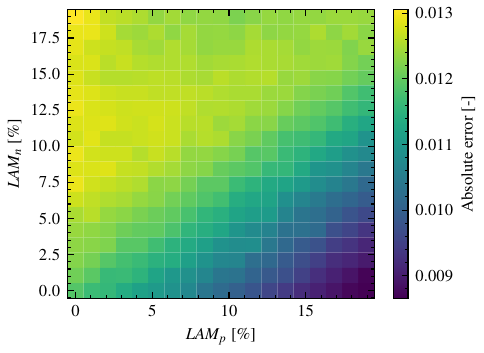}
    \caption{Fine-tuning sweep results across 400 study cases, showing the absolute error in estimated $\epsilon_p$ and $\epsilon_n$ for different levels of Loss of Active Material (LAM).}
    \label{fig: barrido}
\end{figure}

Considering the only varying parameter in $Q_i$ is the $\epsilon$, the change on this parameter will cause an inversely proportional change in $LAM$. This way, we explore 20 cases in a range 0-20\% $LAM$ for both electrodes, resulting in 400 study cases. From the starting point of the Phase 1 pre-training, we fine-tune the PINN for each of these cases, iteration the model until a convergence is reached for both $\epsilon_i$ values. This is achieved by using an early stopping technique over the sum of changes in both $\epsilon_i$, finishing the fine-tuning if a change lower than 1e-7 is produced. The converged parameters are then compared with their expected value by obtaining an absolute error metric that combines both errors by computing their magnitude $AE = \sqrt{\left(\epsilon_p - \hat{\epsilon_p}\right)^2 + \left(\epsilon_n - \hat{\epsilon_n}\right)^2}$. The sweep results are illustrated in Figure \ref{fig: barrido}.

From both Figure \ref{fig: analytic} and the extended analysis from Figure \ref{fig: barrido}, we can observe that the PINNs are able to approximate the unseen active material volume fractions with just the reference voltage. The maximum absolute error magnitude is seen when we have a 0\% $LAM_p$ and 20\% of $LAM_n$ with a 0.01305 or a relative error of 2.19\% (0.92 \% of relative error estimating $\epsilon_p$ and 1.47\% estimating $\epsilon_n$). Greater errors with $LAM_n$ are caused by the worse training performance during the training. Despite this, the error does not deviate significantly and decreases over the rest of the study cases. Also maintain terms related to the PDE, BCs, and BCc lower during the fine-tuning, demonstrating that the PINNs do still capture the electrochemical behaviour appropriately after the parameter update. On average, the local machine required 42.51 ms per iteration (85.02 seconds for 2000), lower than the pre-training process due to the reduced number of parameters to learn. Therefore, the system could be suitable for edge hardware implementation, as will be shown below.


To further validate the effectiveness of PINN-based parameter estimation, we compare its performance with a classical optimization approach applied to the SPM. The optimization is conducted using methods available under the \textit{scipy.optimize} package framework, leveraging gradient-based methods such as BFGS and constrained least-squares fitting to minimize the discrepancy between simulated and reference voltage profiles. This way, we evaluate all suitable methods where no gradient pre-computations are required. The cost function is defined just with the RMSE of the reference and predicted voltages as in Eq \ref{eq.Vloss}. Like with the PINN setup, voltage references are generated in PyBaMM by modifying electrochemical parameters from literature-defined ones and simulating the resulting voltage response. As for the simulated response, the same PyBaMM SPM is programmed into the cost function of the optimizer, where the selected electrochemical parameters are changed by the optimization process.

To enhance the optimization space and reflect the PINN’s ability to handle multiple parameters simultaneously, we extend the classical optimization beyond the ageing-related $\epsilon$ parameters for each electrode to include diffusivity coefficients $D_i$. While we are not able to evaluate this decision against experimental cells, prior observations have demonstrated that PINNs exhibit improved convergence and robustness when optimizing over a wider parameter set. In the classical framework, the additional parameters introduce complexity due to the increased likelihood of local minima. The classical optimization runs until a convergence threshold in parameter updates is met, while we leave an increased 10k iterations for the PINNs to ensure convergence.

By selecting an extreme case for a $\epsilon$ and $D$ variation, we stress-test the ability of both methods to recover meaningful parameters under significant ageing conditions. In this case, we factor $\epsilon_p$ by 1.1, $\epsilon_n$ by 0.75 (allowing an analysis on an extreme case but convergence with PyBaMM), and both diffusivities $D_p$ and $D_n$ by 100. A fine-tuning has been done, setting the learning rate for both $\epsilon_i$ as $\eta_{lr,\epsilon_i} = 0.001\cdot\epsilon_i$ and the diffusion related ones as $\eta_{lr,D_i} = 20\cdot D_i$. The comparative analysis, as shown in Table \ref{tab: classicalvsPINN}, illustrates the improved adaptability of PINN fine-tuning, particularly in its ability to track non-linear parameter evolution. While the classical optimizers can achieve convergence with low computational cost, the resulting values fall far from the expected parameters, remaining as the initial ones or values with high errors. This reinforces the advantage of PINNs in handling high-dimensional, coupled electrochemical systems. We thus prove the practical applicability of physics-informed learning in battery parameter estimation, particularly in scenarios where experimental data is limited or costly to acquire.

\begin{table*}
\centering
\caption{Comparison of PINN-based parameter estimation and classical optimization methods.}
\renewcommand{\arraystretch}{1.2}
\setlength{\tabcolsep}{8pt}
\begin{tabular}{|c|c|c|c|c|c|c|c|c|c|}
\hline
\textbf{Method} & \multicolumn{2}{c|}{\textbf{$\epsilon_p$}} & \multicolumn{2}{c|}{\textbf{$\epsilon_n$}} & \multicolumn{2}{c|}{\textbf{$D_p$}} & \multicolumn{2}{c|}{\textbf{$D_n$}} & \textbf{$t$} \\ 
\cline{2-9}
 & Value & \% & Value & \% & Value & \% & Value & \% & (s) \\ 
\hline
\textit{Initial} & 0.665 & - & 0.75 & - & 4.00E-15 & - & 3.30E-14 & - & -\\ 
\textit{Target} & 0.7315 & - & 0.5625 & - & 4.00E-13 & - & 3.30E-12 & - & -\\ 
\hline
L-BFGS-B & 0.6595 & 9.84 & 0.7734 & 37.49 & 9.12E-16 & 99.77 & 7.52E-15 & 99.77 & 59.46 \\ 
Nelder-Mead & 0.7313 & 0.03 & \textbf{0.5631} & \textbf{0.10} & 1.87E-13 & 53.24 & 1.48E-11 & 342.11 & 164.13 \\ 
Powell & 1.0475 & 43.20 & 4.6495 & 726.58 & 3.82E-11 & 9449.15 & 3.82E-11 & 1057.47 & 12.71 \\ 
\cline{2-9}
trust-constr & \multicolumn{8}{c|}{FAILED} & 49.799 \\ 
COBYLA & \multicolumn{8}{c|}{FAILED} & 8.226 \\ 
\cline{2-9}
TNC & 0.6802 & 7.01 & 0.7118 & 26.55 & 1.50E-15 & 99.62 & 3.25E-14 & 99.02 & 62.73 \\ 
PINN & \textbf{0.7313} & \textbf{0.03} & 0.5606 & 0.34 & \textbf{3.42E-13} & \textbf{14.51} & \textbf{3.58E-12} & \textbf{8.61} & 596.98 \\
\hline
\end{tabular}
\label{tab: classicalvsPINN}
\end{table*}

\section{Experimental validation} \label{Sec:Experimental}

To validate the PINN-based parameter estimation in real conditions, we applied our approach to a degraded M50 lithium-ion cell at 82.09\% of its nominal capacity, implemented on a Raspberry Pi 5 edge device. We can obtain the $LAM$ from both electrodes analysing the open circuit voltage and using the degradation mode estimation tool presented in \cite{Sergi}. The estimation results using this tool are disclosed in the supplementary material S3. This way, we estimate the reference values of the active material volume fractions (\(\epsilon_p, \epsilon_n\)), as well as the stoichiometric limits (\(\theta_{0,p}, \theta_{100,p}, \theta_{0,n}, \theta_{100,n}\)). Considering the cell's electrochemical parameters may vary from literature, we also compute the initial value of the aforementioned parameters from a fresh cell of the same model. Particularly, values displayed in Table \ref{tab: experimental_eps} indicate the initial value in $\epsilon_p$ and stoichiometric limits deviate significantly from the literature parameters. With the initial behaviour corrected, we estimate the aged cell's $\epsilon_p$ and $\epsilon_n$ for a 5.87\% $LAM_p$ and 26.05\% $LAM_n$. Additionally, we have observed discrepancies from literature diffusivities to the real cell at beginning of life. In response, a correction of this parameter is also shown in S3.

\begin{table}
\centering
\caption{Estimated initial and final parameters for the studied M50 cell.}
\renewcommand{\arraystretch}{1.2}
\setlength{\tabcolsep}{8pt}
\begin{tabular}{|c|c|c|}
\hline
\textbf{Parameter} & \textbf{Initial (Fresh Cell)} & \textbf{Final (Degraded Cell)} \\  
\hline
$\epsilon_p$ & 0.575  & 0.5412  \\  
$\epsilon_n$ & 0.774 & 0.5723  \\  
$\theta_{0,p}$ & 0.9458 & 0.8599 \\  
$\theta_{100,p}$ & 0.2715 & 0.2719\\  
$\theta_{0,n}$ & 0.0312 & 0.0339 \\  
$\theta_{100,n}$ & 0.8781 & 0.974 \\  
\hline
\end{tabular}
\label{tab: experimental_eps}
\end{table}

We pre-trained the PINN model using the initial estimated parameters and fine-tuned it with voltage measurements obtained from the degraded cell, under the same conditions and hyperparameters as the analytic case model. Phase 2 is also performed on a similar fashion, assuming stoichiometric parameters do not significantly change and having both $\epsilon_i$ as learned parameters. To avoid discrepancies caused by high C rates from the SPM model to the real cell, we set the charge rate for this phase at 0.3C, the nominal charge rate for this cell. The model is pre-trained in our local machine, and deployed on the Raspberry Pi 5 for the fine-tuning. We acquire the voltage and current data during the mentioned charge condition from the aged cell at controlled 25ºC.

\begin{figure}
    \centering
    \subfloat[\centering\label{fig: eps_exp}]{{\includegraphics[width=0.45\textwidth]{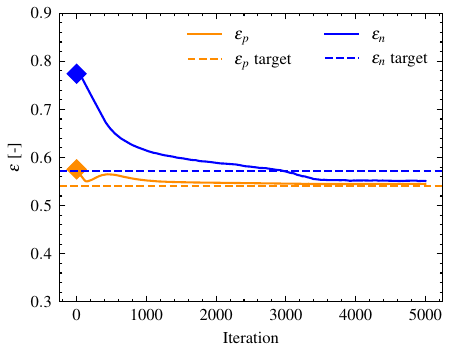}}}\qquad
    \subfloat[\centering\label{fig: V_exp}]{{\includegraphics[width=0.45\textwidth]{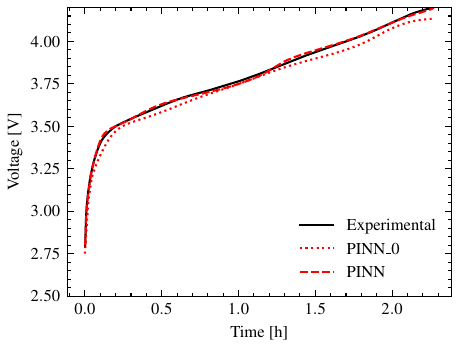}}}\\
    \caption{(a) Evolution of the estimated $\epsilon_p$ and $\epsilon_n$ values during the fine-tuning process on the Raspberry Pi 5, showing convergence towards the expected values. (b) Voltage correction during the fine-tuning.}
    \label{fig: Exp_val}
\end{figure}

The update procedure, presented in Figure \ref{fig: Exp_val}, compares the estimated parameters against the expected $\epsilon_i$ values, demonstrating the ability of the PINN to track the ageing-induced changes. Here, up to 5000 iterations have been allowed for a proper convergence. The estimated $\epsilon_p$ value was 0.545, a relative error of 0.64\%, while we estimated 0.572 for $\epsilon_n$, a 3.84\% relative error to the target value. This higher error for the negative is caused primarily by the change in $\theta_{100,n}$ from the fresh cell to the aged cell. The issue can be solved by adding the stoichiometric parameters to the fine-tuning process, at the cost of a more complex hyperparameter adjusting. We can also observe that the corrected voltages match worse than the analytic scenarios, as more internal parameters and dynamics are not contemplated. Despite this, error magnitude is relatively low, below 3.89\%, demonstrating its effectiveness in real battery degradation scenarios. 

The fine-tuning process has been completed successfully on the potential edge device, allowing periodic updates of the internal state of the cell with high accuracy. It is worth mentioning that the computational cost has increased from the local machine, to 106.1 ms per iteration (for a total of 8 minutes and 51 seconds), and could increase in case more parameters are contemplated to ensure convergence.







\section{Conclusions} \label{Sec:Conclusions}


This study has demonstrated the effectiveness of the proposed PINN approach combined with TL for on-site estimation of lithium-ion battery electrochemical parameters. The two-phase methodology has been successfully validated in both simulated and experimental conditions, confirming that the model is not only functional but also highly efficient in its implementation. The updating methodology has been tested in a wide range of ageing conditions analytically, and validated for a considerably aged battery, proving its robustness. The model only requires few instances from field data and an initial parametrization. In our study case, a nominal charge profile small adjustments from literature parameters were enough to obtain accurate active material volume fraction values from the experimental cell, while performing this estimation on an edge Raspberry Pi 5.


For future work, we believe a holistic approach for parameter estimation is possible using this methodology by contemplating additional electrochemical parameters in the fine-tuning process. The extension of the trainable electrochemical parameters requires adjusting a larger set of hyperparameters for the fine-tuning and added complexity to ensure convergence. Some of the adjustable parameters are not highly observable for certain current input profile conditions. Therefore, a sensitivity analysis among the current profile and each updated parameter is required for this task, as well as including realistic current profiles in the training procedure. In the same way, a more faithful representation of the cell will be made integrating other internal dynamics, like the electrolyte behaviour.




\bibliographystyle{unsrt} 
\bibliography{biblio}

\end{document}